\documentclass[10pt,twocolumn,letterpaper]{article}

\usepackage{cvpr}

\usepackage{subcaption}
\IfFileExists{adjustbox.sty}{\usepackage{adjustbox}}{\newcommand{\adjustbox}[2]{##2}}
\IfFileExists{multirow.sty}{\usepackage{multirow}}{\newcommand{\multirow}[3]{##3}}
\usepackage{booktabs}
\usepackage{tabularx}
\IfFileExists{stfloats.sty}{\usepackage{stfloats}}{}
\usepackage{amsthm}
\usepackage{mathtools}

\newcolumntype{C}{>{\centering\arraybackslash}X}

\usepackage{colortbl}
\usepackage{xcolor}
\usepackage{pgf}
\definecolor{lightred}{rgb}{1,0.8,0.8}   
\definecolor{lightgreen}{rgb}{0.8,1,0.8} 
\newcommand{\rankcolorb}[3]{%
    \pgfmathparse{100*(#3-#2)/(#1-#2)}%
    \xdef\shadeval{\pgfmathresult}%
    \cellcolor{lightgreen!\shadeval!lightred} #3
}

\newcommand{\rankcolors}[3]{%
    \pgfmathparse{100*(#3-#2)/(#1-#2)}%
    \xdef\shadeval{\pgfmathresult}%
    \cellcolor{lightred!\shadeval!lightgreen} #3
}

\newcommand{\projectname}[0]{SeeGroup}

\definecolor{cvprblue}{rgb}{0.21,0.49,0.74}
\usepackage[pagebackref,breaklinks,colorlinks,allcolors=cvprblue]{hyperref}

\title{\projectname{}: Multi-Layer Depth Estimation of Transparent Surfaces\\ via Self-Determined Grouping}

\author{
Hongyu Wen \quad \quad \quad \quad \quad Jia Deng \\
Department of Computer Science, Princeton University \\
{\tt\small \{hongyu.wen,jiadeng\}@princeton.edu}}

\begin{document}
\maketitle

\begin{abstract}

Transparent objects are common in daily life, and it is important to understand their multilayer depth, including the transparent surface and the objects behind it.
Existing methods for multilayer depth typically extend single-layer prediction. They define layers by the front-to-back ordering of 3D points and predict the layers sequentially. However, as layered geometry can admit multiple valid groupings of 3D points into layers, a predefined grouping strategy is inherently restrictive.
In this work, we propose \projectname{}, a multi-layer depth estimation method that avoids imposing a predefined grouping and allows the model itself to adaptively assign surfaces to depth maps. We formulate per-pixel multi-layer depth as a point process, treating depth layers as unordered events along each camera ray. This induces a permutation-invariant likelihood over the observed depth layers, yielding a loss that naturally supports arbitrary layer groupings.
Experiments demonstrate that our method significantly advances the state of the art of multi-layer depth estimation, improving quadruplet relative depth accuracy on LayeredDepth benchmark from 61.34\% to 70.09\%. Code is available at \url{https://github.com/princeton-vl/SeeGroup}.

\end{abstract}

\section{Introduction}
\label{sec:intro}

Transparent objects are common in daily life, and accurate depth estimation for these objects is important for autonomous navigation, 3D reconstruction, and dexterous manipulation.
Although depth estimation has been studied for many years and current methods \cite{metric3dv2, depthanythingv2, depthpro, moge2} achieve strong results on standard benchmarks, they still struggle with transparent objects. One major challenge is that multiple surfaces at different depth can be visible simultaneously, such as a transparent surface and the background objects behind it. As a result, one pixel can correspond to multiple depth values along the same camera ray.

Existing approaches handle this challenge in two main ways. A common choice is to treat transparent objects as if they were opaque and to take only the first depth layer (the transparent surface) as the target geometry \cite{phocal, clearpose, transcg, seeingglass, booster, glasswall}. Another line of work intentionally ignores the transparent surface and focuses solely on predicting the geometry in the background \cite{seeing_through_the_glass, qiu2023looking}. Both ways are limited. In practice, perception systems often need to see both the transparent surface and the background simultaneously. For example, a robot retrieving an item from a plastic container must detect the geometry of the container to avoid collisions while also seeing the target object inside it; similarly, a robot navigating a space with glass doors and walls must localize the glass for safe motion and perceive the scene beyond it to plan a path.

These use cases motivate the study of the \textit{multi-layer depth estimation} task \cite{layereddepth}, where the goal is to predict the depth of all visible surfaces, including possibly multiple transparent surfaces. Given a single RGB image as input, the model should output a sequence of increasing depth values for each pixel, where the length of the sequence is the number of layers and varies by scene and by pixel. To this end, a natural approach is to try to transfer the capabilities of existing single-layer depth estimators to the multi-layer setting, enabling them to predict multiple depth maps.

\begin{figure}
    \centering
    \includegraphics[width=\linewidth]{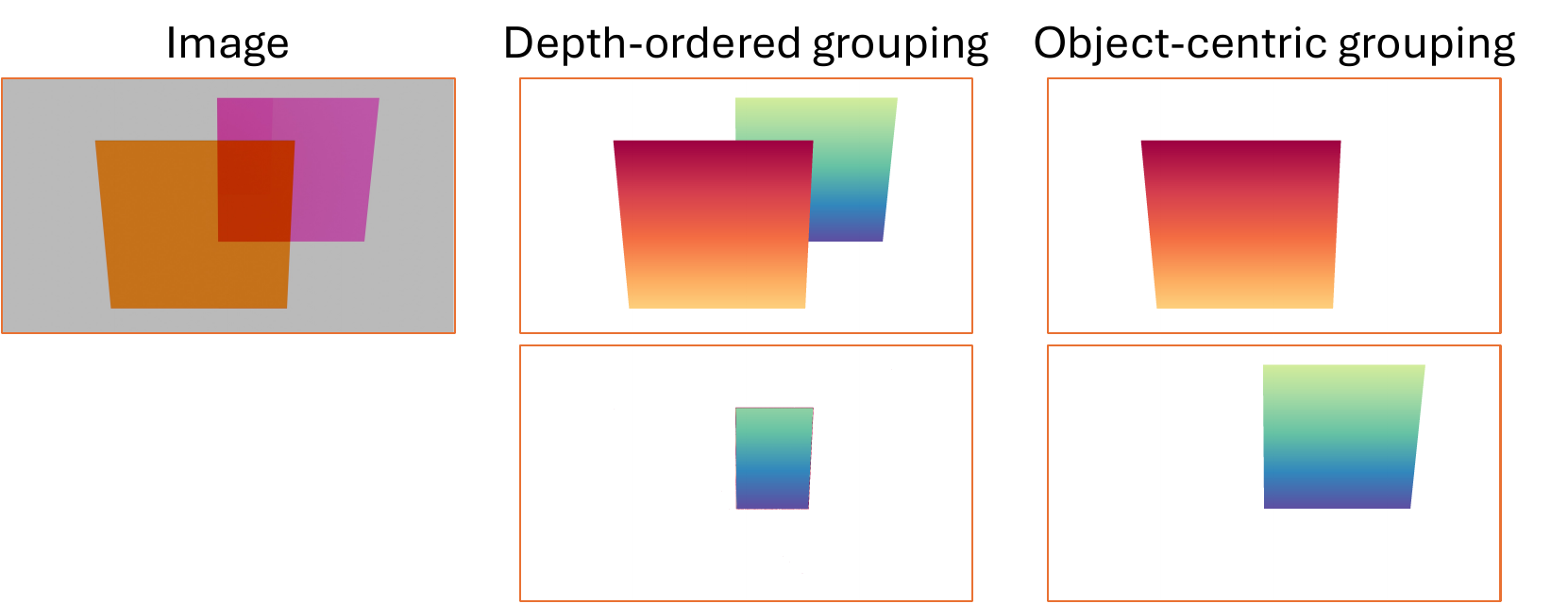}
    \caption{Two example grouping strategies to group multi-layer depth into several depth maps. The best grouping strategy is highly scene-dependent and may very by regions of the image. }
    \label{fig:grouping}
\end{figure}

This immediately raises a design question: How should the model group the sequences of depth values across pixels into a sequence of depth maps? A simple approach proposed by \cite{layereddepth} adopts a depth-ordered strategy, in which the $i$-th map collects the $i$-th depth value at every pixel (\cref{fig:grouping}, middle). This has been shown to work well in many cases, for example, a foreground glass window with a natural background, but it may not be universally optimal and can often produce irregular structures in more complex cases. Consider two partially overlapping transparent planes. Under depth-ordered grouping, the first map combines the entire front plane with the non-overlapping region of the rear plane, and the second map contains only the overlapping region of the rear plane. This causes abrupt changes in both geometry and semantics within each map. In this scenario, an object-centric grouping strategy (\cref{fig:grouping}, right), which assigns the depth values of different objects to different depth maps, yields a more coherent structure and aligns better with human perception.

Intuitively, the best way to group the depth values appears highly scene-dependent and may vary by regions of the image, which means the grouping strategy should not be pre-defined and should be adapted on the fly to the input. Based on this intuition, we propose a novel method, \projectname{}, which allows the model to \textbf{se}lf-d\textbf{e}termine the \textbf{group}ing of the scene across multiple depth layers. In other words, for each pixel, the model predicts a sequence of depth values that are not required to be sorted from low to high; instead, the depth values can be predicted under an arbitrary order decided by the model at inference time. For evaluation, the predicted values can still be sorted into an increasing sequence as needed. The key to enabling this self-determined grouping is a novel loss function that is invariant to the permutation of depth values of each pixel.    

To achieve this, we model the per-pixel multi-layer depths as a point process on the depth axis, where multiple layers along the ray are treated as random events along the depth axis. The point process is characterized by an intensity function, and its value represents the expected rate of surface occurrences at certain depth. Under this formulation, the likelihood of a set of ground-truth depths is proportional to the product of the intensity evaluated at those depths. This product is commutative and permutation invariant with the order of the depths. We parameterize this intensity function as a max-mixture of Laplace components.

\begin{figure*}[t]
    \centering
    \begin{subfigure}[t]{0.76\textwidth}
        \centering
        \adjustbox{valign=t}{\includegraphics[width=\linewidth]{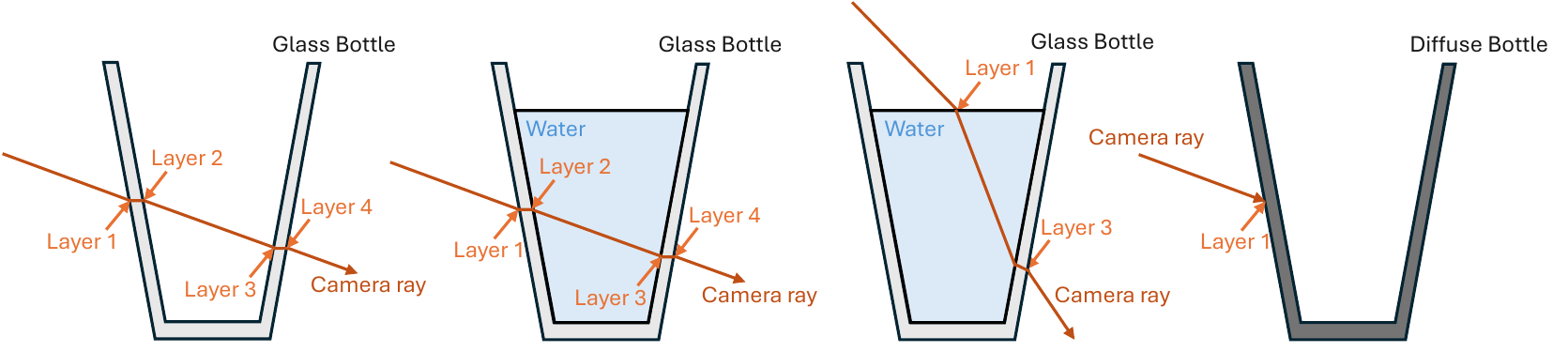}}
        \caption{}
        \label{fig:first}
    \end{subfigure}
    \hfill
    \begin{subfigure}[t]{0.22\textwidth}
        \centering
        \adjustbox{valign=t}{\includegraphics[width=\linewidth]{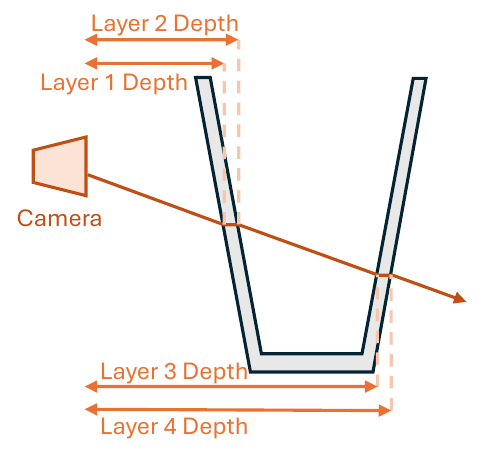}}
        \caption{}
        \label{fig:second}
    \end{subfigure}
    \caption{Definition of multi-layer depth. Figure reproduced from \cite{layereddepth}. (a) Each transition in medium along the camera ray defines a distinct layer. (b) Depth on $i$-th layer is the distance along the z-axis from the $i$-th layer to the camera.}
    \label{fig:def_layer}
\end{figure*}

The Laplace components of the intensity function are predicted by a recurrent decomposition module. Starting from a feature map extracted by a backbone encoder, this module iteratively separates a sequence of feature components from the feature map, where each component captures a dominant group of depth layers. These components are then mapped to the center and scale parameters of Laplace distributions, which define pixel-wise intensity functions over depth. The final depth output at each pixel is obtained from the local maximum of these intensity functions.

In summary, we introduce a novel recurrent decomposition module design and an point process formulation for multi-layer depth. To our knowledge, we are the first to formulate the problem of multi-layer depth grouping and address it with self-determined strategy. Experiments demonstrate that this design significantly improves the state of the art of multi-layer depth estimation. On the LayeredDepth benchmark \cite{layereddepth}, our method achieves the overall best performance across all metrics, improving quadruplet relative depth accuracy from 61.34\% to 70.09\%.

\section{Related Work}
\label{sec:related_work}

\subsection{Monocular Depth Estimation} 

Monocular depth estimation has advanced rapidly alongside the development of deep learning. Early work \cite{eigen2014depth, eigen2015predicting} demonstrated that convolutional neural networks can be used to predict dense depth from a single image. Subsequent research introduced a range of optimization techniques and structural priors, such as ordinal regression \cite{fu2018deep}, geometric constraints \cite{yin2019enforcing}, planar guidance maps \cite{lee2019big}, adaptive bins \cite{bhat2021adabins}, and conditional random fields \cite{newcrfs}. 
Architecturally, progress has been driven by stronger backbones, such as transformer \cite{aich2021bidirectional, li2023depthformer, yang2021transformer, li2022binsformer} and diffusion-based approaches \cite{geowizard, marigold, tosi2024diffusion, gui2025depthfm}. Data-wise, MiDaS \cite{midas} showed that training on a mixture of heterogeneous datasets yielded cross-domain robustness. Following this, recent depth estimation methods \cite{depthanything, depthanythingv2, metric3d, metric3dv2, depthpro, moge, moge2, gui2025depthfm, unidepth, unidepthv2, zoedepth} often leverage strong pre-trained backbones and large-scale multi-dataset training to achieve impressive zero-shot generalization.

However, despite their strong performance on opaque objects, existing depth estimation methods often struggle with transparent objects, both because transparent regions are visually challenging and the lack of training data.

\subsection{Transparent Objects Perception}
Because of the challenging nature of transparent objects, they have been studied specifically for different tasks, such as segmentation \cite{transcut, trans10k, donthit, richcontext}, image matting \cite{tom-net}, optical flow estimation \cite{layeredflow}, and pose estimation \cite{stereobj1m, keypose, seeingglassware, phocal, clearpose}. 

One major challenge is the lack of reliable real-world depth ground truth for transparent objects.
Existing real-world depth benchmarks \cite{phocal, clearpose, transcg, seeingglass, booster, glasswall} typically paint transparent surfaces with a Lambertian coating to create a diffuse twin, then acquire depth using standard sensors. 
This procedure is labor-intensive and hard to scale, insufficient for model training.
To mitigate the data bottleneck, many work trained with synthetic data. A common approach is to construct 3D scenes with transparent objects and render ground-truth depth \cite{cleargrasp, lit, zhu2021rgb, zhang2024towards, liu2025multi}.
This approach is hard to scale because it requires intensive manual labor, and often cannot offer sufficient data for model training. To address this, one common paradigm is to use synthetic data to train depth models. Most build a synthetic scene and render images \cite{cleargrasp, lit, zhu2021rgb, zhang2024towards, liu2025multi, layereddepth}. 
Diffusion models have also been used to generate RGB-depth pairs for transparent objects \cite{tosi2024diffusion}.
\cite{depth4tom} preprocesses inputs by masking transparent regions and replacing them with a uniform color before depth estimation.
\cite{sun2024diffusion} leverage pretrained generative models to inpaint depth in transparent regions.
Another line of work \cite{seeing_through_the_glass, qiu2023looking} focuses on predicting the geometry behind glass, which is typically a simple planar surface.

\begin{figure*}
  \centering
  \includegraphics[width=\textwidth]{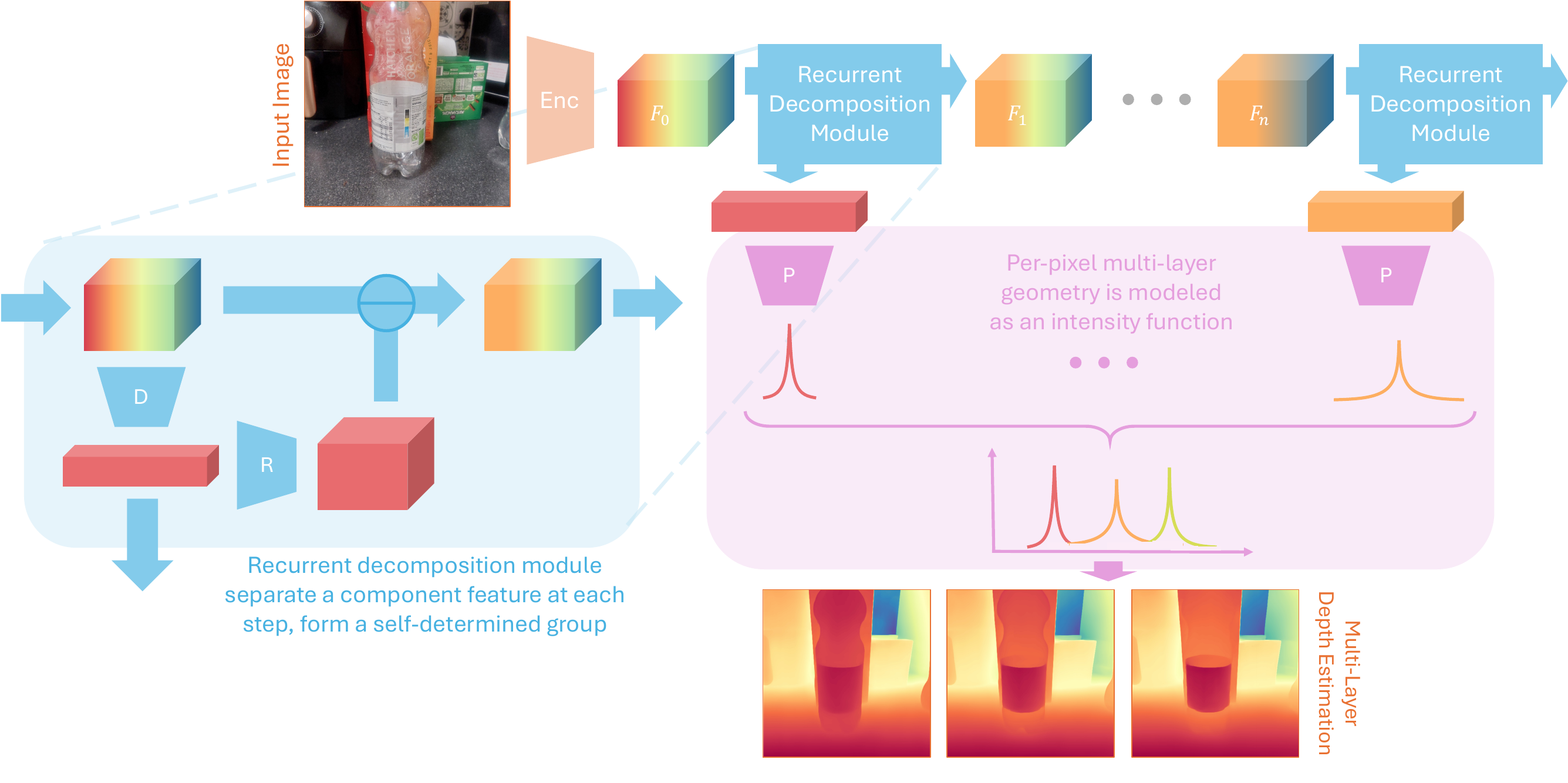}
  \caption{The overall pipeline of \projectname{}. Starting from a feature map extracted by a backbone encoder, a recurrent decomposition module generate a sequence of self-determined feature components. Then these components are mapped to an intensity function over depth, which is parameterized as a max mixture of Laplace functions.}
  \label{fig:method}
\end{figure*}

While these works make progress on depth estimation for transparent objects, they remain limited to single-layer depth prediction. As a result, transparent surfaces, where a single pixel may correspond to multiple depths along the ray, remain inherently ambiguous for these models. In contrast, our method predicts multi-layer depth, enabling a more comprehensive understanding of transparent objects.

\section{Method}
\label{sec:method}

In this section, we first formalize the multi-layer depth estimation task, then present our method, \projectname{}, summarized in \cref{fig:method}.
Our method first uses a recurrent decomposition module to generate a sequence of self-determined feature components, then maps these components to an intensity function over depth, parameterized as a max mixture of Laplace functions. The intensity function will be optimized over permutation invariant loss to keep it order-agnostic.

\subsection{Multi-Layer Depth}
\label{subsec:task}

We follow the definition of multi-layer depth estimation task from LayeredDepth~\cite{layereddepth}. In images with transparent objects, a single camera ray can intersect several transparent surfaces at different depths, and each intersection contributes to the pixel's intensity.
For example, in an image with a glass pane in front of a natural background, a pixel on the pane mixes the texture of the glass pane itself and the background behind it. 
Humans naturally perceive these surfaces and separate them into multiple depth \emph{layers}. Specifically, each pixel corresponds to a camera ray, and every transition between physical media along that ray (for instance, air to glass or air to water) defines a distinct \emph{layer}. Some examples are shown in \cref{fig:first}.

Given an image $\mathcal{I}$ of resolution $H\times W$ and a query pixel $p=(x,y)$, the goal of multi-layer depth estimation is to predict an ordered sequence of per-layer depths $\mathcal{D}(p)=\{d_1,d_2,\dots,d_m\}$, where the number of layers $m$ may vary by pixel. Here $d_i$ is the distance from the camera to the $i$-th layer measured along the optical axis (the $z$-direction), as shown in \cref{fig:second}.

\subsection{Recurrent Decomposition Module}
\label{subsec:recurrent_decomposition}

In this section, we introduce our model architecture design. 

As discussed in \cref{subsec:task}, a single pixel in a region containing transparent objects receives contributions from multiple depth layers, which means $\mathbf{F}_0$ is a mixture of features from multiple surfaces on different depth layers. To predict depth for each surface, we aim to separate this mixture into several components, where each component corresponds to a distinct source of contribution. To this end, we propose a recurrent decomposition module that iteratively extract components $\{\mathbf{C}_1,\mathbf{C}_2,\dots,\mathbf{C}_n\}$ from the initial feature map $\mathbf{F}_0$, with each $\mathbf{C}_i$ representing a salient component of the feature map. The way the mixture is separated is fully learned and self-determined by the model, and the order of these components is irrelevant, since our loss function is permutation invariant.

The recurrent decomposition module consists of two parts: a decomposer ${D}$ and a remapper ${R}$. 
At step $i$, the decomposer takes the current residual feature map $\mathbf{F}_{i-1}$ from the previous step and extracts a component $\mathbf{C}_i = {D}(\mathbf{F}_{i-1})$. Intuitively, $\mathbf{C}_i$ isolates information of the most dominant group of depth layers in current feature map. This component is later used to predict the depth map.

After that, information already explained by $\mathbf{C}_i$ should be removed from the feature map. The remapper projects the component $\mathbf{C}_i$ back to the feature space, producing $\mathbf{F}^{\prime}_{i-1} = {R}(\mathbf{C}_i)$, which represents the portion of $\mathbf{F}_{i-1}$ accounted in the current step. We update $\mathbf{F}_{i}$ by subtracting the remapped features $\mathbf{F}^{\prime}_{i-1}$ to directly remove its contribution from the feature map and encourage the model to focus on the remaining parts.

Putting it all together, the overall pipeline is as follows:
\begin{align*}
    \mathbf{C}_i &= {D}(\mathbf{F}_{i-1}) \\
    \mathbf{F}_{i} &= \mathbf{F}_{i-1} - \eta_i \cdot{R}(\mathbf{C}_i) 
\end{align*}
where $\eta_i$ is a rescaling factor $\eta = \frac{\|\mathbf{F}_{i-1}\|_2}{\|\mathbf{F}^{\prime}_{i-1}\|_2}$ to ensure the scale of $\mathbf{F}_{i}$ remains comparable to $\mathbf{F}_{i-1}$.

\subsection{Intensity Function Parameterization}
\label{subsec:intensity_function_parameterization}

Suppose we have a sequence of component feature maps $\{\mathbf{C}_1,\dots,\mathbf{C}_n\}$, each capturing a distinct source of contribution. Directly regressing multi-layer depths from these components remains challenging, because the contribution of transparent surfaces to the visual features can be weak and noisy. In such regions, solely using an L1 loss may push the model to resolve the ambiguity by overfitting dataset priors rather than relying on actual visual evidence.

To mitigate this, we propose to predict a depth intensity function instead of the depth values. Rather than outputting a sequence of $\{\hat{d}_1, \hat{d}_2, \ldots, \hat{d}_m\}$, we model the per-pixel multi-layer geometry as a point process on the depth axis, where multiple layers of depth along the ray are treated as random events along the ray. The point process is characterized by an intensity function $\mathbf{\Lambda}: \mathbb{R}^{+} \rightarrow \mathbb{R}_{\geq 0}$ defined over the depth axis, which represents the expected rate of surface occurrences. For a small depth interval $[x, x+\mathrm{d} x]$, $\mathbf{\Lambda}(x) \mathrm{d}x$ approximates the expected number of depth layers in that interval. Larger values therefore indicate a higher expected number of surfaces near depth $x$, while values near zero indicate negligible expected occurrence at that depth. This representation explicitly encodes uncertainty: ambiguous regions are represented by low or broad intensity profiles rather than forcing a single discrete depth prediction.

We parameterize $\mathbf{\Lambda}$ as a max-mixture of Laplace components.
A predictor ${P}$ maps each component feature map $\mathbf{C}_i$ to the parameters $d_i \in {\mathbb{R}^+}^{H \times W}$ and $b_i \in {\mathbb{R}^+}^{H \times W}$, where $d_i$ is the center and $b_i$ is the scale of a Laplace component:
\begin{align*}
  \mathbf{L}_i(x) = \frac{1}{2b_i} \exp\left(-\frac{|x - d_i|}{b_i}\right) 
\end{align*}
We then define the overall intensity as
\begin{align*}
  \mathbf{\Lambda} = \max_{i=1}^n \mathbf{L}_i 
\end{align*}
Compare to the standard weighted mixture of Laplace $\sum_{i=1}^n w_i\mathbf{L}_i, \sum_{i=1}^n w_i = 1$, the max-mixture $\max_{i=1}^n \mathbf{L}_i$ is better suited to the multi-layer depth estimation task. At each depth $x$, only the dominant component contributes to $\mathbf{\Lambda}(x)$. This locally suppresses secondary components, encourages different components to specialize in separate depth ranges and reduces the tendency for them to collapse into a single broad mode.

Visually, the per-pixel predicted depth distribution $\mathbf{P}$ forms a curve with multiple peaks, where each peak corresponds to a distinct depth layer. This distributional representation explicitly captures uncertainty over multiple plausible layers, naturally accommodates weak or noisy image evidence, and provides an order-agnostic training target. At inference time, we obtain the multi-layer depth predictions by extracting the peaks of this distribution.

\subsection{Permutation Invariant Loss Function}
\label{subsec:loss}

Note that the intensity function $\boldsymbol{\Lambda}$ can be viewed as a generalization of a probability density: its integral over the depth axis equals the expected number of layers, rather than 1. Under this view, the likelihood of observing ground-truth depths $\left\{d_1, \ldots, d_m\right\}$ at pixel $(x, y)$ is, up to a normalization constant \cite{pointprocess},
\begin{align*}
\mathcal{L}_{(x, y)}\left(\left\{d_i\right\}\right) \propto \prod_{i=1}^m \mathbf{\Lambda}_{(x, y)}\left(d_i\right)
\end{align*}

As multiplication of probabilities is commutative, this objective is inherently permutation invariant. As a result, maximizing this likelihood does not impose any fixed ordering of depth layers: the model is free to determine the order of depths at each pixel and to learn its own grouping of depths into the output maps.

For the numerical stability, the objective is in log space. The training objective of intensity function is given by:
\begin{align*}
  {\mathcal{L}_{int}}_{(x, y)}&(\{d_1, d_2, \ldots, d_m\}) \\
  &= -\log \mathbf{\Lambda}_{(x, y)}(\{d_1, d_2, \ldots, d_m\}) \\
  &= -\log \prod_{i=1}^m \mathbf{\Lambda}_{(x, y)}(d_i) \\
  &= -\sum_{i=1}^m \log \mathbf{\Lambda}_{(x, y)}(d_i) \\
  &= -\sum_{i=1}^m \log \max_{j=1}^n \mathbf{L}_j(d_i) 
\end{align*}

\begin{figure*}
  \centering
  \includegraphics[width=\linewidth]{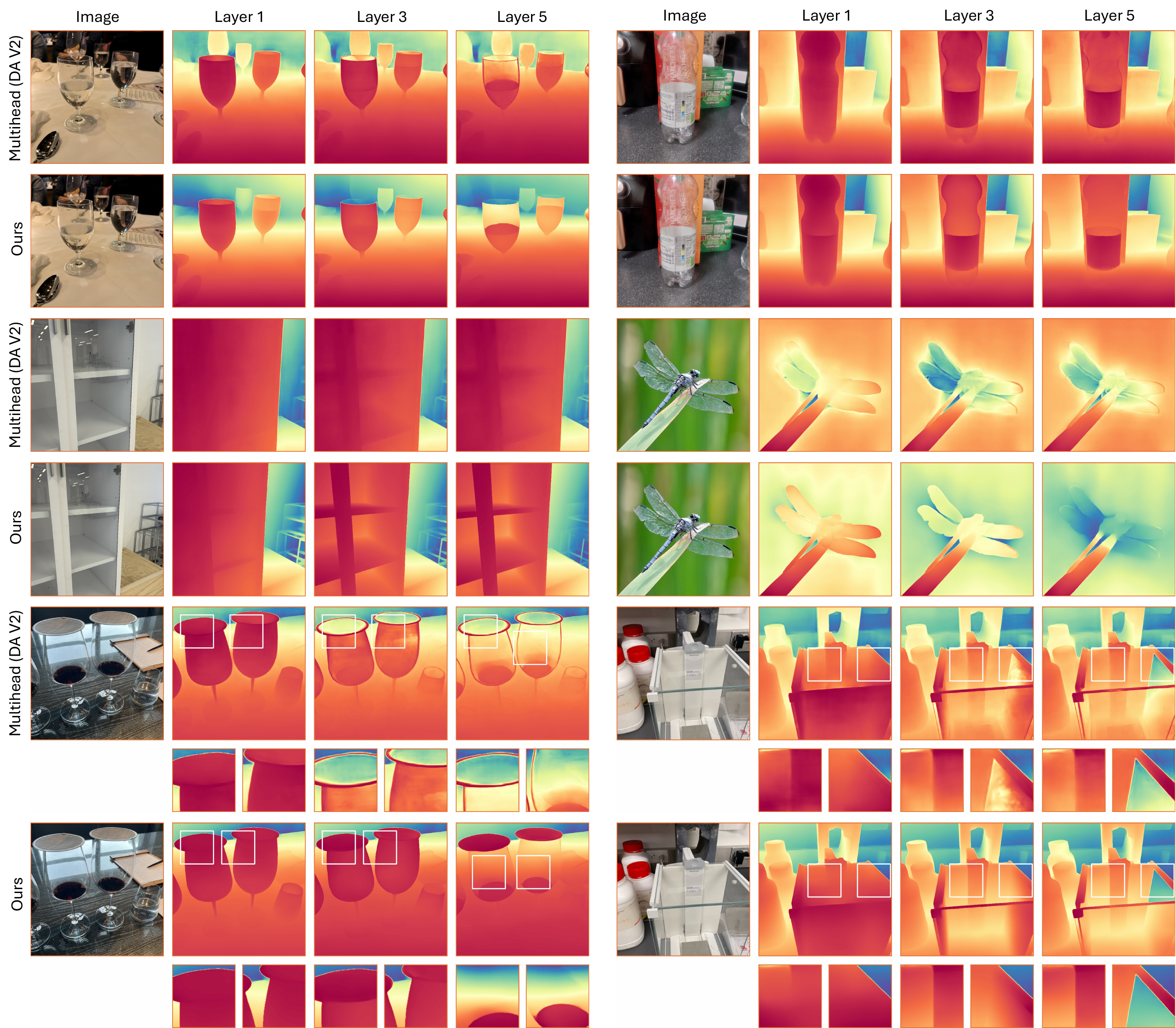}
  \caption{Qualitative results on LayeredDepth benchmark. Our method produce sharper results with less artifacts.}
  \label{fig:qualitative_results_real}
\end{figure*}

\begin{table*}[t!]
  \centering
  \begingroup
    \newcommand{\tub}[1]{\underline{\textbf{#1}}}
    \newcommand{\tb}[1]{\textbf{#1}}
    \newcommand{\tu}[1]{\underline{#1}}
  \resizebox{\linewidth}{!}{
\begin{tabular}{l c ccc c ccc c ccc c ccc c ccc}
    \toprule
        \multirow{2}{*}{Method} && 
        \multicolumn{3}{c}{All} && 
        \multicolumn{3}{c}{Mixed} && 
        \multicolumn{3}{c}{Layer 1} && 
        \multicolumn{3}{c}{Layer 3} && 
        \multicolumn{3}{c}{Layer 5} \\
        && Q & T & P && Q & T & P && Q & T & P
        && Q & T & P && Q & T & P \\
    \midrule
Multi-head (NeWCRFs)
&& \rankcolorb{70.09}{23.77}{25.32} & \rankcolorb{74.88}{40.70}{41.65} & \rankcolorb{82.62}{62.26}{63.95}
&& \rankcolorb{70.84}{22.97}{25.44} & \rankcolorb{76.94}{42.01}{46.22} & \rankcolorb{92.56}{69.40}{74.91}
&& \rankcolorb{72.73}{24.05}{24.16} & \rankcolorb{77.64}{39.85}{39.85} & \rankcolorb{87.04}{65.59}{65.59}
&& \rankcolorb{63.75}{20.50}{27.60} & \rankcolorb{66.19}{30.42}{38.39} & \rankcolorb{72.87}{48.59}{54.22}
&& \rankcolorb{60.21}{20.34}{27.11} & \rankcolorb{62.32}{27.73}{34.28} & \rankcolorb{64.32}{42.41}{47.33} \\

Index Concat (NeWCRFs)
&& \rankcolorb{70.09}{23.77}{26.90} & \rankcolorb{74.88}{40.70}{45.25} & \rankcolorb{82.62}{62.26}{66.35}
&& \rankcolorb{70.84}{22.97}{28.67} & \rankcolorb{76.94}{42.01}{52.45} & \rankcolorb{92.56}{69.40}{80.74}
&& \rankcolorb{72.73}{24.05}{24.05} & \rankcolorb{77.64}{39.85}{41.42} & \rankcolorb{87.04}{65.59}{66.72}
&& \rankcolorb{63.75}{20.50}{30.01} & \rankcolorb{66.19}{30.42}{41.61} & \rankcolorb{72.87}{48.59}{55.66}
&& \rankcolorb{60.21}{20.34}{29.86} & \rankcolorb{62.32}{27.73}{40.34} & \rankcolorb{64.32}{42.41}{52.26} \\

Recurrent (NeWCRFs)
&& \rankcolorb{70.09}{23.77}{23.77} & \rankcolorb{74.88}{40.70}{40.70} & \rankcolorb{82.62}{62.26}{62.26}
&& \rankcolorb{70.84}{22.97}{22.97} & \rankcolorb{76.94}{42.01}{42.01} & \rankcolorb{92.56}{69.40}{69.40}
&& \rankcolorb{72.73}{24.05}{25.91} & \rankcolorb{77.64}{39.85}{44.64} & \rankcolorb{87.04}{65.59}{68.13}
&& \rankcolorb{63.75}{20.50}{20.50} & \rankcolorb{66.19}{30.42}{30.42} & \rankcolorb{72.87}{48.59}{48.59}
&& \rankcolorb{60.21}{20.34}{20.34} & \rankcolorb{62.32}{27.73}{27.73} & \rankcolorb{64.32}{42.41}{42.41} \\

Multi-head (DA v2)
&& \underline{\rankcolorb{70.09}{23.77}{61.34}} & \underline{\rankcolorb{74.88}{40.70}{70.57}} & \underline{\rankcolorb{82.62}{62.26}{82.56}}
&& \underline{\rankcolorb{70.84}{22.97}{60.23}} & \underline{\rankcolorb{76.94}{42.01}{71.91}} & \underline{\rankcolorb{92.56}{69.40}{90.41}}
&& \underline{\rankcolorb{72.73}{24.05}{65.62}} & \underline{\rankcolorb{77.64}{39.85}{74.41}} & \underline{\rankcolorb{87.04}{65.59}{85.81}}
&& \underline{\rankcolorb{63.75}{20.50}{54.36}} & \underline{\rankcolorb{66.19}{30.42}{60.75}} & \textbf{\rankcolorb{72.87}{48.59}{72.87}}
&& \underline{\rankcolorb{60.21}{20.34}{51.99}} & \underline{\rankcolorb{62.32}{27.73}{56.42}} & \underline{\rankcolorb{64.32}{42.41}{63.99}} \\

    \midrule

    \projectname{} (Ours)
&& \textbf{\rankcolorb{70.09}{23.77}{70.09}} & \textbf{\rankcolorb{74.88}{40.70}{74.88}} & \textbf{\rankcolorb{82.62}{62.26}{82.62}}
&& \textbf{\rankcolorb{70.84}{22.97}{70.84}} & \textbf{\rankcolorb{76.94}{42.01}{76.94}} & \textbf{\rankcolorb{92.56}{69.40}{92.56}}
&& \textbf{\rankcolorb{72.73}{24.05}{72.73}} & \textbf{\rankcolorb{77.64}{39.85}{77.64}} & \textbf{\rankcolorb{87.04}{65.59}{87.04}}
&& \textbf{\rankcolorb{63.75}{20.50}{63.75}} & \textbf{\rankcolorb{66.19}{30.42}{66.19}} & \underline{\rankcolorb{72.87}{48.59}{69.14}}
&& \textbf{\rankcolorb{60.21}{20.34}{60.21}} & \textbf{\rankcolorb{62.32}{27.73}{62.32}} & \textbf{\rankcolorb{64.32}{42.41}{64.32}} \\

    \bottomrule
\end{tabular}
  }
  \caption{Multi-layer depth estimation methods evaluated on LayeredDepth benchmark test set via tuple-wise accuracy. Best scores are in \textbf{bold}. Second best \underline{underlined}.}
  \label{tab:eval_real}
  \endgroup
\end{table*}

However, this objective is one-sided: it encourages the intensity to be high at every ground-truth depth, but it does not penalize the network for predicting extra components that do not correspond to any ground-truth layer. To discourage such over-prediction, we introduce a component-coverage loss that enforces the complementary constraint that each predicted component should be supported by at least one ground-truth depth. This loss is defined as:
\begin{align*}
  {\mathcal{L}_{cov}}_{(x, y)}(\{d_1, d_2, \ldots, d_m\}) = -\sum_{j=1}^n \log \max_{i=1}^m \mathbf{L}_j(d_i) 
\end{align*}
For each component $\mathbf{L}_j$, we evaluate its intensity at all ground-truth depths and keep only the best match via $\max_i \mathbf{L}_j(d_i)$. If $\mathbf{L}_j$ is well aligned with at least one ground-truth depth, this maximum is large and the resulting penalty is small. Conversely, if $\mathbf{L}_j$ is far from all ground-truth depths, $-\log\left(\max_i \mathbf{L}_j(d_i)\right)$ becomes large, penalizing this spurious component.

Combined with the original intensity function term, this loss yields a bidirectional matching objective that reduces over-prediction while remaining permutation invariant.

In addition, we also found that the gradient-matching loss $\mathcal{L}_{gm}$ \cite{megadepth} is helpful to improve the fine details of the predicted depths. The overall training objective is given by:
\begin{align*}
  \mathcal{L} = \lambda_{int} \mathcal{L}_{int} + \lambda_{cov} \mathcal{L}_{cov} + \lambda_{gm} \mathcal{L}_{gm} \\
\end{align*}

\section{Experiments}
\label{sec:experiments}

\subsection{Implementation Details}
\projectname{} is implemented in PyTorch \cite{PyTorch} and optimized using AdamW \cite{adamw} optimizer with an initial learning rate of $1\times10^{-5}$. Following \cite{layereddepth}, we initialize the feature extractor from the metric-depth checkpoints of Depth Anything V2 \cite{depthanythingv2}, which are based on DINOv2-ViT-L \cite{dinov2}. The iteration of recurrent decomposition is set to $n = 4$.

We train \projectname{} on the LayeredDepth-Syn dataset \cite{layereddepth}. LayeredDepth-Syn is a procedurally generated synthetic dataset, built on infinigen-indoors \cite{infinigen_indoors}, specifically designed for transparent objects with multi-layer depth annotations. It contains 14,800 training images and 500 validation images. Our full model is trained for 250k steps on $4\times$ L40 GPUs with a batch size of 4. During training, the model is trained in scale-invariant manner following \cite{midas} and $\lambda_{int}, \lambda_{cov}, \lambda_{gm}$ are set to 1.0, 0.1, 1.0, respectively.

At inference time, we extract multi-layer depths from the predicted intensity by detecting peaks along the depth axis. To avoid double-counting, we suppress peaks whose depths differ by less than 0.02. 

\subsection{Evaluation on Real Benchmark}
We zero-shot evaluate our method on LayeredDepth \cite{layereddepth}, a real-world multi-layer depth benchmark consisting of 1,500 images of diverse scenes and 14.2M relative depth tuples. To the best of our knowledge, it is the only real-world benchmark with multi-layer depth annotations. The evaluation metric is tuple-wise accuracy: the percentage of relative depth tuples for which the model predicts the correct ordering, reported separately for pairs (P), triplets (T), and quadruplets (Q). Among these, quadruplet accuracy is the most informative metric, because the chance of being correct by random guessing is smallest.

We report results on six subsets: (i) All: all tuples; (ii) Mixed: tuples whose points lie on different layers; and (iii) Layer $i$: tuples whose points all lie on the same layer $i$. Following \cite{layereddepth}, we report results only for odd-numbered layers ($i \in {1,3,5}$), because in most cases even-numbered layers have depths very close to the preceding odd-numbered layer. We compare against four models proposed in \cite{layereddepth}: Multi-head (NeWCRFs), Index Concat (NeWCRFs), Recurrent (NeWCRFs), and Multi-head (DA v2). Among these, our method shares the same pre-trained single-layer depth backbone as Multi-head (DA v2).

The results on test set are shown in \cref{tab:eval_real}. \projectname{} outperforms all baselines on 14 of the 15 evaluation metrics. In particular, it improves quadruplet accuracy over the Multi-head (DA v2) baseline from 61.34\% to 70.09\%, a 14.26\% relative gain. Notably, \projectname{} achieves these improvements with fewer parameters, demonstrating the effectiveness and efficiency of our design.

Qualitative results are shown in \cref{fig:qualitative_results_real}. Our method produces noticeably sharper predictions with fewer artifacts, and succeeds to recover multiple layers when the baseline model fails. The right panel of the first row particularly highlights the benefit of self-determined grouping: the background food bag is likely treated as a single coherent layer, allowing the model to predict a smooth depth map, even though most of it is occluded by the foreground bottle.

However, as multi-layer depth estimation remains an extremely challenging task and the available data are quite limited, there is considerable room for improvement. A common failure mode is over-prediction, as illustrated in the left panel of the first row, where the back side of the glass bottle is incorrectly split into two depth layers. Another failure mode occurs on highly out-of-distribution examples, as shown in the right panel of the second row. We hope this work will inspire further research in this direction.

\begin{table*}[t]
  \centering
  \begingroup
    \newcommand{\tub}[1]{\underline{\textbf{#1}}}
    \newcommand{\tb}[1]{\textbf{#1}}
    \newcommand{\tu}[1]{\underline{#1}}
  \resizebox{\linewidth}{!}{
\begin{tabular}{l c ccc c ccc c ccc c ccc c ccc}
    \toprule
        \multirow{2}{*}{Method} && 
        \multicolumn{3}{c}{All} && 
        \multicolumn{3}{c}{Mixed} && 
        \multicolumn{3}{c}{Layer 1} && 
        \multicolumn{3}{c}{Layer 3} && 
        \multicolumn{3}{c}{Layer 5} \\
        && Q & T & P && Q & T & P && Q & T & P
        && Q & T & P && Q & T & P \\
    \midrule
    \multicolumn{21}{c}{Model Architecture} \\
    \midrule
    
        Multi-head
&& \textbf{\rankcolorb{73.04}{67.82}{73.04}} & \underline{\rankcolorb{76.54}{72.77}{75.72}} & \underline{\rankcolorb{83.21}{81.70}{82.39}}
&& \underline{\rankcolorb{71.73}{66.18}{70.69}} & \underline{\rankcolorb{78.52}{70.74}{73.08}} & \rankcolorb{92.47}{88.41}{88.41}
&& \textbf{\rankcolorb{78.36}{71.03}{78.36}} & \textbf{\rankcolorb{81.62}{77.49}{81.62}} & \textbf{\rankcolorb{89.60}{87.56}{89.60}}
&& \underline{\rankcolorb{66.84}{64.63}{66.64}} & \underline{\rankcolorb{68.94}{66.14}{68.31}} & \underline{\rankcolorb{70.06}{66.45}{67.06}}
&& \rankcolorb{66.14}{39.64}{39.64} & \rankcolorb{67.65}{41.26}{41.26} & \rankcolorb{67.77}{51.05}{51.05} \\

GRU
&& \rankcolorb{73.04}{67.82}{67.82} & \rankcolorb{76.54}{72.77}{72.77} & \rankcolorb{83.21}{81.70}{81.70}
&& \rankcolorb{71.73}{66.18}{66.18} & \rankcolorb{78.52}{70.74}{70.74} & \underline{\rankcolorb{92.47}{88.41}{88.44}}
&& \rankcolorb{78.36}{71.03}{71.03} & \rankcolorb{81.62}{77.49}{77.49} & \underline{\rankcolorb{89.60}{87.56}{88.15}}
&& \rankcolorb{66.84}{64.63}{64.63} & \rankcolorb{68.94}{66.14}{66.14} & \rankcolorb{70.06}{66.45}{66.45}
&& \underline{\rankcolorb{66.14}{39.64}{44.44}} & \underline{\rankcolorb{67.65}{41.26}{47.48}} & \underline{\rankcolorb{67.77}{51.05}{55.89}} \\

RD (Ours)
&& \underline{\rankcolorb{73.04}{67.82}{71.50}} & \textbf{\rankcolorb{76.54}{72.77}{76.54}} & \textbf{\rankcolorb{83.21}{81.70}{83.21}}
&& \textbf{\rankcolorb{71.73}{66.18}{71.73}} & \textbf{\rankcolorb{78.52}{70.74}{78.52}} & \textbf{\rankcolorb{92.47}{88.41}{92.47}}
&& \underline{\rankcolorb{78.36}{71.03}{73.60}} & \underline{\rankcolorb{81.62}{77.49}{78.71}} & \rankcolorb{89.60}{87.56}{87.56}
&& \textbf{\rankcolorb{66.84}{64.63}{66.84}} & \textbf{\rankcolorb{68.94}{66.14}{68.94}} & \textbf{\rankcolorb{70.06}{66.45}{70.06}}
&& \textbf{\rankcolorb{66.14}{39.64}{66.14}} & \textbf{\rankcolorb{67.65}{41.26}{67.65}} & \textbf{\rankcolorb{67.77}{51.05}{67.77}} \\

    \midrule
    \multicolumn{21}{c}{Intensity Parameterization} \\
    \midrule

Weighted Mixture
&& \underline{\rankcolorb{69.03}{52.68}{56.12}} & \underline{\rankcolorb{72.80}{61.00}{62.02}} & \rankcolorb{79.52}{73.56}{73.56}
&& \underline{\rankcolorb{69.56}{53.59}{61.76}} & \underline{\rankcolorb{74.42}{63.46}{70.14}} & \underline{\rankcolorb{86.26}{82.53}{83.24}}
&& \underline{\rankcolorb{70.30}{50.81}{52.02}} & \rankcolorb{74.34}{57.21}{57.21} & \rankcolorb{83.87}{74.25}{74.25}
&& \underline{\rankcolorb{66.55}{57.94}{59.21}} & \underline{\rankcolorb{67.70}{60.75}{62.41}} & \underline{\rankcolorb{67.74}{65.21}{66.21}}
&& \rankcolorb{55.28}{45.40}{45.40} & \underline{\rankcolorb{56.65}{47.97}{50.49}} & \underline{\rankcolorb{60.37}{55.85}{58.06}} \\

Sorted Mixture
&& \rankcolorb{69.03}{52.68}{52.68} & \rankcolorb{72.80}{61.00}{61.00} & \underline{\rankcolorb{79.52}{73.56}{74.43}}
&& \rankcolorb{69.56}{53.59}{53.59} & \rankcolorb{74.42}{63.46}{63.46} & \rankcolorb{86.26}{82.53}{82.53}
&& \rankcolorb{70.30}{50.81}{50.81} & \underline{\rankcolorb{74.34}{57.21}{60.18}} & \underline{\rankcolorb{83.87}{74.25}{76.84}}
&& \rankcolorb{66.55}{57.94}{57.94} & \rankcolorb{67.70}{60.75}{60.75} & \rankcolorb{67.74}{65.21}{65.21}
&& \underline{\rankcolorb{55.28}{45.40}{45.54}} & \rankcolorb{56.65}{47.97}{47.97} & \rankcolorb{60.37}{55.85}{55.85} \\

Max-Mixture (Ours)
&& \textbf{\rankcolorb{69.03}{52.68}{69.03}} & \textbf{\rankcolorb{72.80}{61.00}{72.80}} & \textbf{\rankcolorb{79.52}{73.56}{79.52}}
&& \textbf{\rankcolorb{69.56}{53.59}{69.56}} & \textbf{\rankcolorb{74.42}{63.46}{74.42}} & \textbf{\rankcolorb{86.26}{82.53}{86.26}}
&& \textbf{\rankcolorb{70.30}{50.81}{70.30}} & \textbf{\rankcolorb{74.34}{57.21}{74.34}} & \textbf{\rankcolorb{83.87}{74.25}{83.87}}
&& \textbf{\rankcolorb{66.55}{57.94}{66.55}} & \textbf{\rankcolorb{67.70}{60.75}{67.70}} & \textbf{\rankcolorb{67.74}{65.21}{67.74}}
&& \textbf{\rankcolorb{55.28}{45.40}{55.28}} & \textbf{\rankcolorb{56.65}{47.97}{56.65}} & \textbf{\rankcolorb{60.37}{55.85}{60.37}} \\

    \midrule
    \multicolumn{21}{c}{Loss Functions} \\
    \midrule
    SiLog
&& \rankcolorb{69.03}{42.49}{48.20} & \rankcolorb{72.80}{54.65}{60.85} & \rankcolorb{79.52}{71.45}{74.64}
&& \rankcolorb{69.56}{44.77}{46.84} & \rankcolorb{74.42}{59.28}{62.64} & \rankcolorb{86.26}{81.53}{83.73}
&& \rankcolorb{71.22}{40.83}{50.99} & \rankcolorb{74.34}{53.36}{63.81} & \rankcolorb{83.87}{73.30}{79.16}
&& \rankcolorb{66.71}{44.04}{44.26} & \rankcolorb{68.23}{50.81}{50.81} & \rankcolorb{68.29}{60.84}{60.84}
&& \rankcolorb{55.28}{32.27}{32.27} & \rankcolorb{56.65}{39.30}{39.30} & \rankcolorb{60.37}{51.09}{51.09} \\

L1
&& \rankcolorb{69.03}{42.49}{42.49} & \rankcolorb{72.80}{54.65}{54.65} & \rankcolorb{79.52}{71.45}{71.45}
&& \rankcolorb{69.56}{44.77}{44.77} & \rankcolorb{74.42}{59.28}{59.28} & \rankcolorb{86.26}{81.53}{81.53}
&& \rankcolorb{71.22}{40.83}{40.83} & \rankcolorb{74.34}{53.36}{53.36} & \rankcolorb{83.87}{73.30}{73.30}
&& \rankcolorb{66.71}{44.04}{44.04} & \rankcolorb{68.23}{50.81}{51.03} & \rankcolorb{68.29}{60.84}{61.20}
&& \rankcolorb{55.28}{32.27}{36.74} & \rankcolorb{56.65}{39.30}{44.12} & \underline{\rankcolorb{60.37}{51.09}{56.42}} \\

L1 + GM
&& \rankcolorb{69.03}{42.49}{57.00} & \rankcolorb{72.80}{54.65}{64.26} & \rankcolorb{79.52}{71.45}{75.60}
&& \rankcolorb{69.56}{44.77}{57.98} & \rankcolorb{74.42}{59.28}{67.80} & \underline{\rankcolorb{86.26}{81.53}{85.81}}
&& \rankcolorb{71.22}{40.83}{56.48} & \rankcolorb{74.34}{53.36}{63.75} & \rankcolorb{83.87}{73.30}{78.34}
&& \rankcolorb{66.71}{44.04}{57.50} & \rankcolorb{68.23}{50.81}{59.95} & \rankcolorb{68.29}{60.84}{63.83}
&& \underline{\rankcolorb{55.28}{32.27}{49.95}} & \underline{\rankcolorb{56.65}{39.30}{53.62}} & \rankcolorb{60.37}{51.09}{56.39} \\

Int + GM
&& \underline{\rankcolorb{69.03}{42.49}{68.36}} & \underline{\rankcolorb{72.80}{54.65}{72.00}} & \underline{\rankcolorb{79.52}{71.45}{78.90}}
&& \underline{\rankcolorb{69.56}{44.77}{66.58}} & \underline{\rankcolorb{74.42}{59.28}{72.34}} & \rankcolorb{86.26}{81.53}{85.77}
&& \textbf{\rankcolorb{71.22}{40.83}{71.22}} & \underline{\rankcolorb{74.34}{53.36}{74.26}} & \underline{\rankcolorb{83.87}{73.30}{82.87}}
&& \textbf{\rankcolorb{66.71}{44.04}{66.71}} & \textbf{\rankcolorb{68.23}{50.81}{68.23}} & \textbf{\rankcolorb{68.29}{60.84}{68.29}}
&& \rankcolorb{55.28}{32.27}{47.91} & \rankcolorb{56.65}{39.30}{49.38} & \rankcolorb{60.37}{51.09}{55.92} \\

Int + Cov + GM (Ours)
&& \textbf{\rankcolorb{69.03}{42.49}{69.03}} & \textbf{\rankcolorb{72.80}{54.65}{72.80}} & \textbf{\rankcolorb{79.52}{71.45}{79.52}}
&& \textbf{\rankcolorb{69.56}{44.77}{69.56}} & \textbf{\rankcolorb{74.42}{59.28}{74.42}} & \textbf{\rankcolorb{86.26}{81.53}{86.26}}
&& \underline{\rankcolorb{71.22}{40.83}{70.30}} & \textbf{\rankcolorb{74.34}{53.36}{74.34}} & \textbf{\rankcolorb{83.87}{73.30}{83.87}}
&& \underline{\rankcolorb{66.71}{44.04}{66.55}} & \underline{\rankcolorb{68.23}{50.81}{67.70}} & \underline{\rankcolorb{68.29}{60.84}{67.74}}
&& \textbf{\rankcolorb{55.28}{32.27}{55.28}} & \textbf{\rankcolorb{56.65}{39.30}{56.65}} & \textbf{\rankcolorb{60.37}{51.09}{60.37}} \\
    \bottomrule
\end{tabular}
  }
  \caption{Ablation results on LayeredDepth  validation set via tuple-wise accuracy. Best scores are in \textbf{bold}. Second best \underline{underlined}.}
  \label{tab:ablation}
  \endgroup
\end{table*}

\begin{table*}[t!]
  \centering
  \resizebox{\linewidth}{!}{
    \begin{tabularx}{1.3\textwidth}{l CCCC CCCC CCCC CCCC}
        \toprule
            \multirow{2}{*}{Method} 
            & \multicolumn{4}{c}{Layer 1} & \multicolumn{4}{c}{Layer 3}
            & \multicolumn{4}{c}{Layer 5} \\
            & AbsRel$\downarrow$ & RMS$\downarrow$ & $\delta$1$\uparrow$ & $\delta$2$\uparrow$
            & AbsRel$\downarrow$ & RMS$\downarrow$ & $\delta$1$\uparrow$ & $\delta$2$\uparrow$ 
            & AbsRel$\downarrow$ & RMS$\downarrow$ & $\delta$1$\uparrow$ & $\delta$2$\uparrow$ \\
    \toprule
Multi-head (NeWCRFs)
& \rankcolors{17.97}{8.24}{17.97} & \rankcolors{25.53}{15.21}{23.41} & \rankcolorb{93.43}{81.89}{83.08} & \rankcolorb{97.73}{93.14}{93.14}
& \rankcolors{17.61}{10.68}{15.90} & \rankcolors{51.79}{24.42}{47.39} & \rankcolorb{88.92}{76.39}{80.15} & \rankcolorb{96.83}{92.53}{93.89}
& \rankcolors{17.25}{11.15}{14.58} & \rankcolors{52.53}{28.37}{47.51} & \rankcolorb{87.92}{76.54}{81.48} & \rankcolorb{96.40}{92.20}{94.12} \\

Index Concat (NeWCRFs)
& \rankcolors{17.97}{8.24}{17.26} & \rankcolors{25.53}{15.21}{23.70} & \rankcolorb{93.43}{81.89}{83.02} & \rankcolorb{97.73}{93.14}{93.27}
& \rankcolors{17.61}{10.68}{16.25} & \rankcolors{51.79}{24.42}{48.19} & \rankcolorb{88.92}{76.39}{79.63} & \rankcolorb{96.83}{92.53}{93.72}
& \rankcolors{17.25}{11.15}{15.03} & \rankcolors{52.53}{28.37}{48.11} & \rankcolorb{87.92}{76.54}{80.61} & \rankcolorb{96.40}{92.20}{93.97} \\

Recurrent (NeWCRFs)
& \rankcolors{17.97}{8.24}{17.23} & \rankcolors{25.53}{15.21}{25.53} & \rankcolorb{93.43}{81.89}{81.89} & \rankcolorb{97.73}{93.14}{93.37}
& \rankcolors{17.61}{10.68}{17.61} & \rankcolors{51.79}{24.42}{51.79} & \rankcolorb{88.92}{76.39}{76.39} & \rankcolorb{96.83}{92.53}{92.53}
& \rankcolors{17.25}{11.15}{17.25} & \rankcolors{52.53}{28.37}{52.53} & \rankcolorb{87.92}{76.54}{76.54} & \rankcolorb{96.40}{92.20}{92.20} \\

Multi-head (DA v2)
& \textbf{\rankcolors{17.97}{8.24}{8.24}} & \textbf{\rankcolors{25.53}{15.21}{15.21}} & \textbf{\rankcolorb{93.43}{81.89}{93.43}} & \textbf{\rankcolorb{97.73}{93.14}{97.73}}
& \textbf{\rankcolors{17.61}{10.68}{10.68}} & \underline{\rankcolors{51.79}{24.42}{37.97}} & \underline{\rankcolorb{88.92}{76.39}{88.76}} & \textbf{\rankcolorb{96.83}{92.53}{96.83}}
& \textbf{\rankcolors{17.25}{11.15}{11.15}} & \underline{\rankcolors{52.53}{28.37}{40.67}} & \underline{\rankcolorb{87.92}{76.54}{87.51}} & \textbf{\rankcolorb{96.40}{92.20}{96.40}} \\

\projectname{} (Ours)
& \underline{\rankcolors{17.97}{8.24}{9.82}} & \underline{\rankcolors{25.53}{15.21}{17.00}} & \underline{\rankcolorb{93.43}{81.89}{90.72}} & \underline{\rankcolorb{97.73}{93.14}{97.09}}
& \underline{\rankcolors{17.61}{10.68}{11.70}} & \textbf{\rankcolors{51.79}{24.42}{24.42}} & \textbf{\rankcolorb{88.92}{76.39}{88.92}} & \underline{\rankcolorb{96.83}{92.53}{96.15}}
& \underline{\rankcolors{17.25}{11.15}{13.29}} & \textbf{\rankcolors{52.53}{28.37}{28.37}} & \textbf{\rankcolorb{87.92}{76.54}{87.92}} & \underline{\rankcolorb{96.40}{92.20}{95.30}} \\

    \bottomrule
    \end{tabularx}
  }
  \caption{Multi-layer depth estimation methods evaluated on our LayeredDepth-Syn validation set. 
  Values are scaled by 100 for clearer comparison.
  Best scores are in \textbf{bold}. Second best \underline{underlined}.
  }
  \label{tab:eval_syn}
\end{table*}

\subsection{Ablation Study}
We conduct ablation studies along three axes: (a) model architecture, (b) intensity parameterization, and (c) loss functions. The results are summarized in \cref{tab:ablation}. For the model architecture variants, all models use the same encoder as the full model, and we report results on the LayeredDepth validation split. For the intensity parameterization and loss function variants, we use DINOv2-ViT-S as the feature encoder and evaluate on the LayeredDepth validation split, due to computational constraints.

\begin{table}[t!]
  \centering
  \resizebox{\linewidth}{!}{
    \begin{tabularx}{0.6\textwidth}{l CCCCC}
        \toprule
            {Method} 
            & \#Param & \#Param w/o enc & CPU time (s) & GPU time (s) & FLOPs (T) \\
    \toprule
Multi-head & 428M & 124M & 24.31 & 10.02 & 1.13 \\
GRU       & 731M & 427M & 23.84 & 15.59 & 2.44 \\
RD (ours) & 356M &  51M & 16.29 & 9.41 & 1.23\\
    \bottomrule
    \end{tabularx}
  }
  \caption{We report the total number of parameters (\#Param), the number of parameters excluding the pretrained encoder (\#Param w/o enc), CPU time and GPU time per forward pass, and FLOPs for all three architectures. Our Recurrent Decomposition Module (RD) uses the fewest parameters among them.}
  \label{tab:params}
\end{table}

\begin{figure*}
  \centering
  \includegraphics[width=\linewidth]{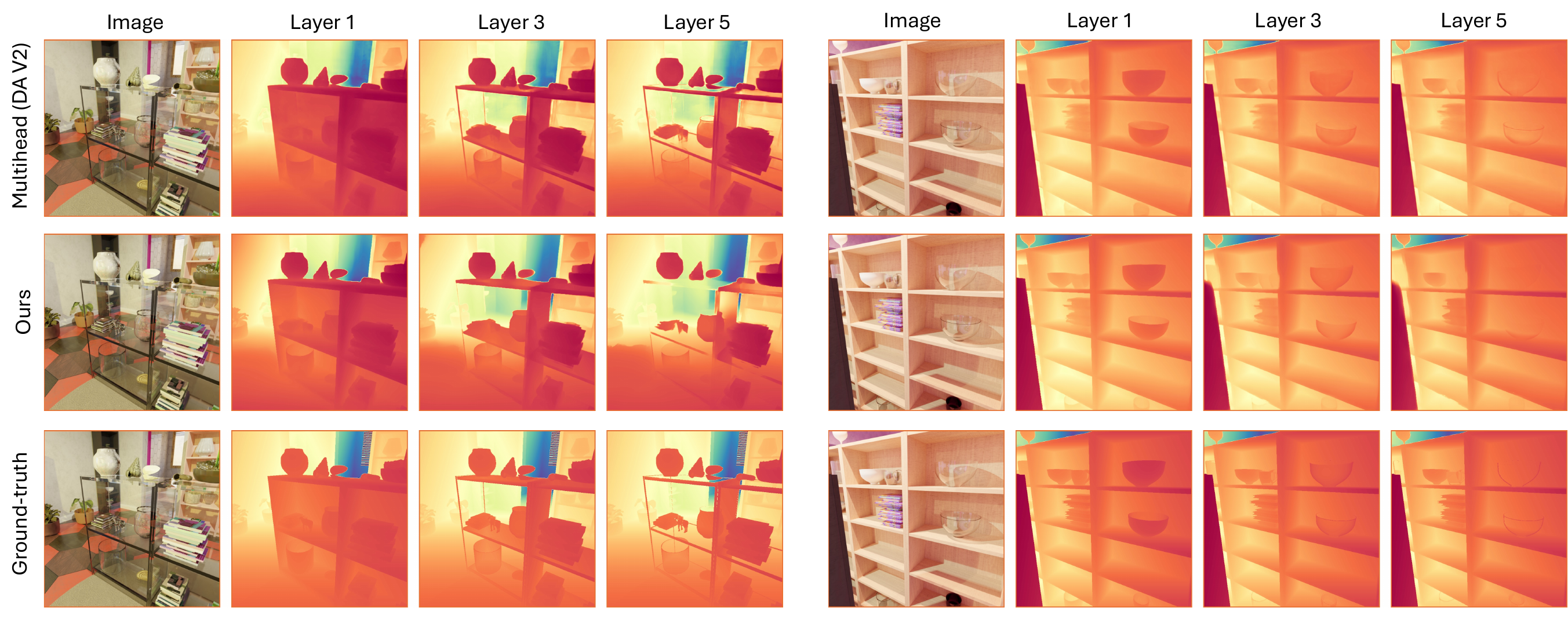}
  \caption{Qualitative results on LayeredDepth-Syn validation set. Our model produces smooth, coherent background layers. }
  \label{fig:qualitative_results_syn}
\end{figure*}

\paragraph{Model Architecture.}
To evaluate the effectiveness of our recurrent decomposition module (RD), we compare it against two baseline architectures:
\begin{itemize}
\item \textbf{Multi-head}: We replicate the Depth Anything V2 depth head $n$ times, with each head responsible for predicting one Laplace component.
\item \textbf{GRU}: We use a GRU \cite{gru} to iteratively update the image feature map $\mathbf{F}_i$, and at each step a shared depth head predicts one Laplace component.
\end{itemize}
We also compare the total number of parameters, overall runtime, and FLOPs of all three models, as shown in \cref{tab:params}. Our model achieves the best overall performance while using the fewest parameters, with the only exception being Layer 1, where it is outperformed by the Multi-head variant. However, likely because it lacks information exchange across layers, the Multi-head design degrades catastrophically on deeper layers.

\paragraph{Intensity Parameterization.}
We evaluate the effectiveness of different intensity parameterization design.
\begin{itemize}
    \item \textbf{Weighted Mixture.} We model the intensity as a uniformly weighted mixture of Laplace components $\frac{1}{n}\sum_{i=1}^n w_i\mathbf{L}_i$.
    \item \textbf{Ordered Distribution.} Instead of modeling an order-agnostic intensity over depths, we assign
    a single Laplace probability distribution to each layer. This corresponds to depth-ordered grouping and is not permutation invariant.
\end{itemize}

The max-mixture parameterization achieves the best performance on most subsets. It is slightly worse than the sorted-mixture baseline on the first layer, but substantially better on the remaining layers, which confirms the effectiveness of our self-determined grouping and permutation-invariant loss design.

\paragraph{Loss Functions.}
We also evaluate the effectiveness of our loss design by comparing it with SiLog \cite{silog_loss} and L1 loss. Our design achieves the best overall performance.

\subsection{Evaluation on Synthetic Data}

To further quantify performance, we also evaluate on the LayeredDepth-Syn validation set. However, this evaluation is less informative than the real-world benchmark, because (1) all existing multi-layer depth estimation models are trained on LayeredDepth-Syn, so this is an in-domain evaluation rather than a zero-shot setting, and (2) the dataset has limited diversity, making it more susceptible to overfitting.
We use the relative point error (AbsRel), root mean square error (RMS) and the percentage of inlier $\delta_i$, $i \in \{ 1, 2\} $ with threshold $1.25^i$ as metrics. The prediction is aligned with the ground truth using an estimated global shift and scale.

The results are shown in \cref{tab:eval_syn}. Our method shows slightly worse results than Multi-head (DAv2) but still significantly better than NeWCRFs-based results. 

Our method achieves better results on the real-world benchmark but slightly worse results on the synthetic dataset, possibly for two reasons. First, our model has fewer parameters, which may make it harder to fit a limited synthetic data distribution. Second, the training data contain noise: the LayeredDepth-Syn ground truth is ray-tracer-based and often produces noisy boundaries around transparent objects. Since our method decomposes the feature representation and assigns each component to a different depth range, it tends to learn smooth, consistent background layers rather than overfitting to these noisy boundaries, as illustrated in \cref{fig:qualitative_results_syn}.

\section{Conclusion}
\label{conclusion}

We introduce \projectname{}, a new method for multi-layer depth estimation that combines a recurrent decomposition module with an point process formulation of multi-layer depth. \projectname{} significantly advances the state of the art in multi-layer depth estimation and is broadly useful for transparent object understanding.

\section*{Acknowledgments} This work was partially supported by the National Science Foundation.

\section{Appendix}

\subsection{Additional Details}

\subsubsection{Model Design}
In a forward pass, we first extract features $\mathbf{F}_0$ from the input RGB image $\mathcal{I} \in \mathbb{R}^{H \times W \times 3}$ using a DINOv2 \cite{dinov2} backbone, following the pipeline of Depth Anything V2 \cite{depthanythingv2}. Specifically, $\mathbf{F}_0$ consists of a sequence of feature maps with shape $(B, N_0, H_0, W_0)$, where $B$ is the batch size, $N_0$ is the channel dimension, and $H_0, W_0$ denote the spatial resolution in terms of patch counts along height and width.
These feature maps are extracted from intermediate layers of the backbone. As a result, each map corresponds to a different stage of the network and captures information at a distinct scale, ranging from coarse to fine. In our implementation, the sequence length is $l = 4$.

The architecture of our recurrent decomposition module follows the overall design of Depth Anything V2 \cite{depthanythingv2}. For the decomposer $D$, we first apply a $1 \times 1$ convolution to each backbone feature map, obtaining tensors of shape $(B, N_{1,i}, H_0, W_0)$, where $N_{1,i}$ depends on the backbone stage $i$. These tensors are then passed through either a ConvTranspose layer or a Conv layer (depending on the desired resolution) to produce feature maps of shape $(B, N_{1,i}, H_i, W_i)$. Here $(H_i, W_i)$ and $N_{1,i}$ vary with $i$, with coarser stages generally having greater spatial resolution and fewer channels.
Each of these feature maps is then mapped by another convolutional layer to a shared channel dimension, resulting in tensors of shape $(B, N_2, H_i, W_i)$. The resulting sequence of multi-scale feature maps at step $k$ forms $\mathbf{C}_k$ in the recurrent decomposition.

In the predictor $P$, the sequence $\mathbf{C}_k$ is first passed through two convolutional layers that align the spatial resolutions and channel dimensions of all feature maps. We then fuse these maps in a top-down manner, from the finest to the coarsest level, by combining adjacent pairs. For each pair of adjacent levels, we first upsample the finer-level feature using a convolutional block to match the resolution of the coarser one, sum the two features, and pass the result through another convolutional block. After all levels are fused, we obtain a single aggregated feature map of shape $(B, N_2, 2H_1, 2W_1)$. This feature map is further processed by two convolutional layers to produce the parameters $d_i, b_i \in \mathbb{R}_{>0}^{H \times W}$, where $d_i$ is the center (depth) and $b_i$ is the scale of the Laplace-shaped contribution.

The remapper $R$ is essentially the inverse of the decomposer $D$. Starting from the aggregated feature map, it first uses a convolutional layer to reconstruct, for each scale $i$, a feature map of shape $(B, N_{1,i}, H_i, W_i)$. A second convolutional layer then maps these features to a common resolution $(B, N_{1,i}, H_0, W_0)$, followed by a final convolution that projects them back to the original backbone feature space $(B, N_0, H_0, W_0)$. The resulting feature map is subtracted from the previous residual feature map $\mathbf{F}_{k-1}$ to obtain the updated feature map $\mathbf{F}_k$.
The exact values of $N_0$, $N_{1,i}$, $N_2$, and other architectural hyperparameters are listed in \cref{tab:architecture}.

\begin{table}[t]
  \centering
  \begingroup
    \newcommand{\tub}[1]{\underline{\textbf{#1}}}
    \newcommand{\tb}[1]{\textbf{#1}}
    \newcommand{\tu}[1]{\underline{#1}}
  \resizebox{\linewidth}{!}{
\begin{tabular}{ccccc}
    \toprule
        $N_0$ & $\{H_i\}$ & $\{W_i\}$ & $\{N_{1,i}\}$ & $N_2$ \\
        1024 & $\{148, 74, 37, 19\}$ & $\{148, 74, 37, 19\}$ & $\{256, 512, 1024, 1024\}$ & 256\\
    \midrule
    \bottomrule
\end{tabular}
  }
  \caption{The exact values of architectural hyperparameters in recurrent decomposition module. }
  \label{tab:architecture}
  \endgroup
\end{table}

\subsubsection{Training}
During training, we follow the scale-invariant normalization of \cite{midas}. Given a depth map $d$, we normalize it to $\tilde{d}$ as
\begin{align*}
t &= \operatorname{median}(d), \\
s &= \operatorname{mean}(|d - t|), \\
\tilde{d} &= \frac{d - t}{s}.
\end{align*}
The scale parameters $b$ of the Laplace distributions are clipped to the range $[1, 10]$. We clip the gradient norm to $0.1$ to stabilize training. We use the AdamW \cite{adamw} optimizer with an initial learning rate of $1 \times 10^{-5}$ for the main network and $1 \times 10^{-6}$ for pretrained the DINOv2 backbone. The momentum parameters are set to $(0.9, 0.99)$, and the weight decay is set to $0.01$. The learning rate multiplier at training step $k$ is given by $\left(1 - \tfrac{k}{\text{total steps}}\right)^{0.9}$. All images are resized to $518 \times 518$ during training. For the gradient matching loss, all layers are weighted by $(1.2, 1.0, 1.0, 1.0)$. During inference, we first resize the image so that its shorter side is 518 pixels while preserving the aspect ratio, and then round the longer side to the nearest multiple of 14. The predicted depth maps are finally bilinearly upsampled back to the original image resolution.

\subsection{Additional Results}
\begin{figure*}
    \centering
    \includegraphics[width=\linewidth]{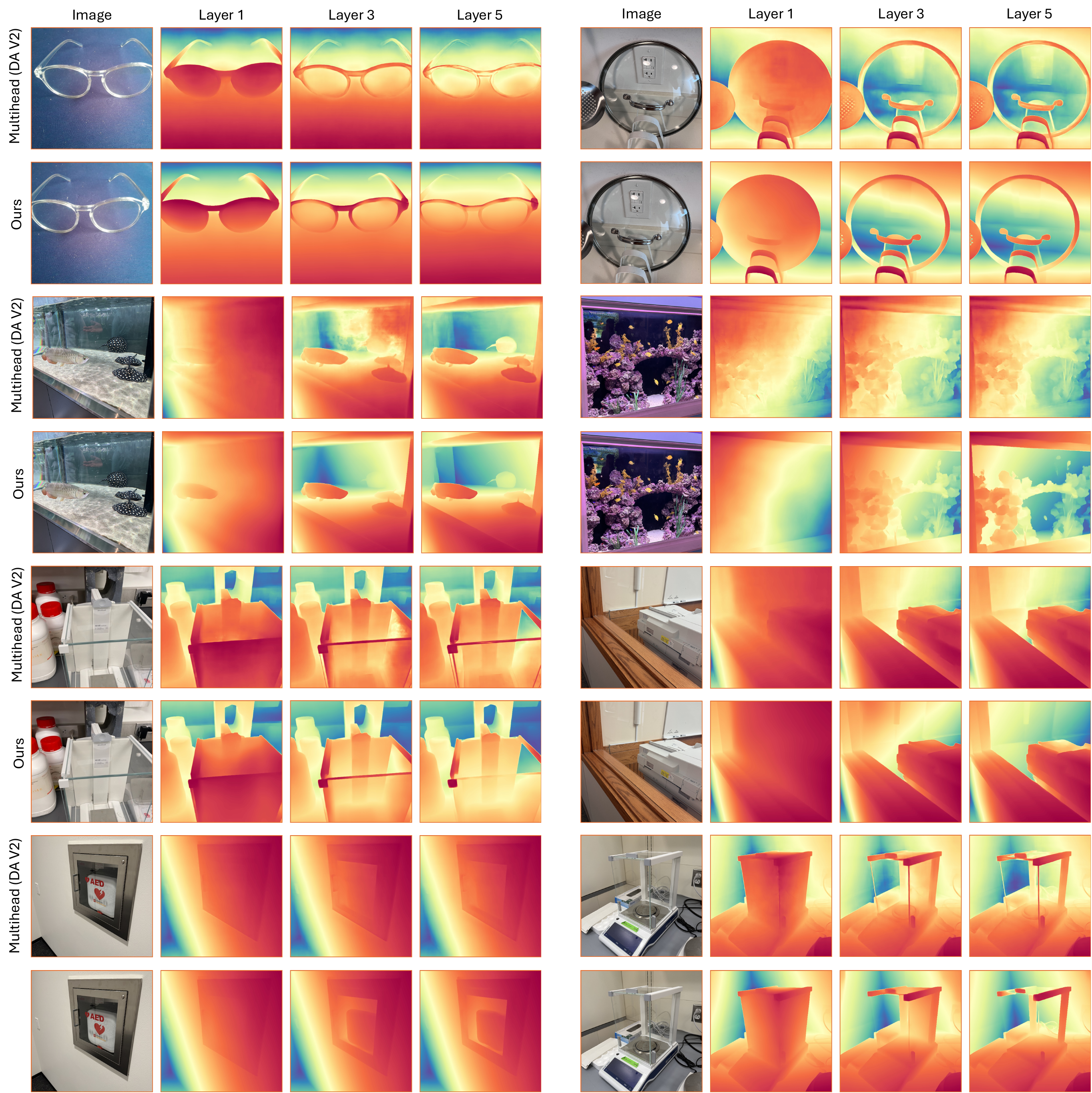}
    \caption{Additional qualitative results on LayeredDepth.}
    \label{fig:additional_quali_real}
\end{figure*}

To help readers better visually assess the effectiveness of our datasets, we provide additional qualitative results on real world benchmark LayeredDepth \cite{layereddepth}, as shown in \cref{fig:additional_quali_real}. Our method produces noticeably sharper predictions with fewer artifacts.

{
    \small
    \bibliographystyle{ieeenat_fullname}
    \bibliography{main}
}

\end{document}